\documentclass[10pt,journal,compsoc]{IEEEtran}

\ifCLASSOPTIONcompsoc
  \usepackage[nocompress]{cite}
\else
  \usepackage{cite}
\fi
\usepackage{amssymb}

\ifCLASSINFOpdf
  \usepackage[pdftex]{graphicx}
\else
\fi
\usepackage{amsmath}

\usepackage{xcolor}

\usepackage{hyperref}
\usepackage{pifont}
\usepackage{diagbox}
\usepackage{multirow}
\usepackage{ragged2e}
\usepackage{colortbl}

\newcommand{\cmark}{\ding{51}}%
\newcommand{\xmark}{\ding{55}}%

\newcommand{\etal}{\textit{et al}.}
\newcommand{\ie}{\textit{i}.\textit{e}.}
\newcommand{\eg}{\textit{e}.\textit{g}.}

\DeclareMathOperator*{\argmax}{\arg\!\max}

\begin{document}

\title{Flow-based Spatio-Temporal Structured Prediction of Motion Dynamics}

\author{Mohsen~Zand,~\IEEEmembership{Member,~IEEE,}
        Ali~Etemad,~\IEEEmembership{Senior Member,~IEEE,}
        and~Michael~Greenspan,~\IEEEmembership{Member,~IEEE}
\IEEEcompsocitemizethanks{\IEEEcompsocthanksitem 

Mohsen Zand, Ali Etemad, and Michael Greenspan are with the Department of Electrical and Computer Engineering, 
and Ingenuity Labs Research Institute,Queen's University, Kingston, ON, Canada. \\ 
E-mail: (m.zand, ali.etemad, michael.greenspan)@queensu.ca \\
Corresponding author: Mohsen Zand
}
}

\markboth{IEEE TRANSACTIONS ON PATTERN ANALYSIS AND MACHINE INTELLIGENCE,~Vol.~X, No.~X, X~2023}%
{Zand \MakeLowercase{\textit{et al.}}: Flow-based Spatio-Temporal Structured Prediction of Motion Dynamics}

\IEEEtitleabstractindextext{%
\begin{abstract}
\justifying{Conditional Normalizing Flows (CNFs) are flexible generative models capable of representing complicated distributions with high dimensionality and large interdimensional correlations, making them appealing for structured output learning. Their effectiveness in modelling multivariates spatio-temporal structured data has yet to be completely investigated. We propose MotionFlow as a novel normalizing flows approach that autoregressively conditions the output distributions on the spatio-temporal input features. It combines deterministic and stochastic representations with CNFs to create a probabilistic neural generative approach that can model the variability seen in high-dimensional structured spatio-temporal data. 
We specifically propose to use conditional priors to factorize the latent space for the time dependent modeling. We also exploit the use of masked convolutions as autoregressive conditionals in CNFs. As a result, our method is able to define arbitrarily expressive output probability distributions under temporal dynamics in multivariate prediction tasks. We apply our method to different tasks, including trajectory prediction, motion prediction, time series forecasting, and binary segmentation, and demonstrate that our model is able to leverage normalizing flows to learn complicated time dependent conditional distributions.}

\end{abstract}

}

\maketitle

\IEEEdisplaynontitleabstractindextext

\IEEEpeerreviewmaketitle

\ifCLASSOPTIONcompsoc
\IEEEraisesectionheading{\section{Introduction}\label{sec:introduction}}
\else
\section{Introduction}
\label{sec:introduction}
\fi

\IEEEPARstart{L}{earning} dynamic relations is an active area of research in computer vision due to its useful applications in forecasting complex trajectories. It however remains a challenging task to automate since the underlying relationships or dynamical model evolve over time into a stochastic sequential process with a high degree of inherent uncertainty~\cite{gui2018adversarial, martinez2017human, liu2019towards, gong2021memory}. 

To model dynamic relations among the model components, the complex structured-sequence of spatio-temporal dynamics should be learned. To address this challenge, a variety of state-based methods~\cite{taylor2009factored, lehrmann2014efficient} and deep learning approaches~\cite{fragkiadaki2015recurrent,martinez2017human, kipf2018neural} have been proposed to cast the dynamic relationships features into pseudo images in order to learn the movement patterns. 
Nonetheless, these methods are unable to deal with the large variations of potential outputs, which result in a loss of dynamic diversity across the components (\ie body joints), resulting in over-smoothing 
where the estimated trajectories tend to converge to a mean position over time without clear movements~\cite{li2021skeleton}. 

Most recent methods rely on generative adversarial networks (GANs) or variational autoencoders (VAEs) for learning spatial information, followed by recurrent neural networks (RNNs) for learning temporal information~\cite{barsoum2017hpgan,kundu2019bihmp,aliakbarian2020stochastic,yan2018mt}. Such models however do not perform well with overly complex structures, and are also often hard to train, encounter mode collapse, posterior collapse, or vanishing gradients~\cite{kobyzev2020normalizing}.

Recently, flow-based generative methods have shown promising performance in modeling complex outputs~\cite{papamakarios2017masked, ng2018actionflownet, kingma2018glow,ho2019flow++}. These models are based on normalizing flows, a family of generative models with tractable distributions that consist of sequential invertible components~\cite{kobyzev2020normalizing}. 
Flow-based methods have been successfully applied to non-temporal data such as images~\cite{kingma2016improving,papamakarios2017masked,ziegler2019latent}. We believe their ability to learn multi-output distributions makes them particularly attractive for learning dynamic relationships. 

In this paper, we introduce \emph{MotionFlow}, a flow-based structured prediction model designed to learn spatio-temporal relations in a dynamic systems, which is applicable to applications such as the forecasting of trajectories. Our model learns more robust spatio-temporal representations while maintaining the structure of high-dimensional data. Specifically, MotionFlow is a conditional autoregressive flow-based solution which can directly model the log-likelihood of temporal and spatial information for relatively long sequences. 
To the best of our knowledge, our work is the first attempt to extend normalizing flows for spatio-temporal structured prediction.

We evaluate our method by performing extensive experiments and ablation studies on different datasets, 
and illustrate that MotionFlow outperforms other methods in the field by generating more accurate and consistent predictions both qualitatively and quantitatively.

Our contributions are summarized as follows:
\begin{itemize}
    \item We extend normalizing flows for spatio-temporal density estimation tasks. We exploit the flexibility of conditional normalizing flows and use masked convolutions as autoregressive conditionals. We also propose to use conditional priors to factorize the latent space for the time-dependent modeling. In particular, a novel normalizing flows approach is developed that autoregressively conditions the output distributions on the spatio-temporal input features. 
    
    \item We apply our method to different tasks, including trajectory prediction, motion prediction, time series forecasting, and binary segmentation, and demonstrate that our model is able to leverage normalizing flows to learn complicated time dependent conditional distributions. 
    
    \item We perform extensive experimental evaluations on challenging datasets and show the effectiveness of the proposed method on different tasks, achieving state-of-the-art in most cases. 
    
\end{itemize}
    
We also make our implementation public~\footnote{https://github.com/MohsenZand/MotionFlow} to enable reproducibility and contribute to the field.
In the remainder of this paper, we first review the related literature in Sec.~\ref{sec:related_work}. Some background materials that are essential in our method are included in Sec.~\ref{sec:background}. 
Our proposed MotionFlow is then presented in more detail in Sec.~\ref{sec:proposed}. Finally, in Sec.~\ref{sec:experiments}, we analyze the functionality of the model through extensive experiments and demonstrate how it works in different tasks, and summarize and conclude in Sec.~\ref{sec:conclusion}.

\section{Related Work}\label{sec:related_work}

Our method relates to the topics of structured prediction and flow-based representation learning. We briefly review the most relevant related literature in the following subsections.

\subsection{Structured Prediction}

The goal in structured prediction is to learn a function between the input $x$ and the structured output $y$. For instance, a semantic segmentation map is returned for a given input image. The conditional probability $p(y|x)$ hence captures the relationship between the structured output and the input features. 
A variety of approaches have been studied for modeling structures between output variables~\cite{nowozin2011structured,belanger2016structured, tu2018learning,graber2018deep,lu2020structured}. These methods are mainly a generalization of the standard classification algorithms such as conditional random field (CRF)~\cite{sutton2006introduction}, structured SVM (SSVM)~\cite{tsochantaridis2004support}, and structured perceptron~\cite{collins2002discriminative} to model the correlations between output variables. 

Belanger and McCallum~\cite{belanger2016structured} proposed structured prediction energy networks (SPENs) which used an energy function for scoring structured outputs. Gradient descent was then used to optimize the assigned energies with respect to the ground-truth outputs. Graber~\etal~\cite{graber2018deep} proposed the use of output variables as an intermediate structured layer between deep neural structure layers to capture non-linear interactions among output variables. Graber and Schwing~\cite{graber2019graph} presented graph structured prediction energy networks to better model local and higher-order correlations between output variables. An exact inference for structured prediction is however NP-hard. Thus, different types of prediction regularization~\cite{niculae2018sparsemap} and approximate inference~\cite{hazan2010primal,tu2018learning} have been developed. In~\cite{aksan2019structured}, a structured prediction layer is introduced to the task of 3D human motion modelling. It modeled the structure of the human skeleton and the spatial dependencies between joints.

\subsection{Flow-based Representation}

Normalizing Flows have been successfully explored as a family of generative models with tractable distributions~\cite{atanov2019semi,hoogeboom2019emerging,Kumar2019VideoFlowAF,kobyzev2020normalizing}. A Normalizing Flow is a transformation of a simpler probability distribution into a more complicated distribution by a sequence of invertible and differentiable functions. Many applications based on normalizing flows have recently emerged in the literature, such as density estimation~\cite{dinh2014nice}, variational inference~\cite{rezende2015variational}, image generation~\cite{kingma2018glow}, and noise modelling~\cite{abdelhamed2019noise}. For a comprehensive review of this concept, we refer the reader to~\cite{kobyzev2020normalizing}.

Flow-based generative models have also been investigated for complex outputs. Dinh~\etal~\cite{dinh2016density} proposed real-valued non-volume preserving (real NVP) transformations to model high-dimensional data. They used invertible and learnable transformations for exact sampling, inference of latent variables, and log-density estimation of data samples. Kingma and Dhariwal~\cite{kingma2018glow} proposed `Glow' as a generative flow by leveraging more invertible layers. They used invertible $1\!\times\!1$ convolutions in their model and employed it to synthesize high-resolution natural images. In~\cite{ho2019flow++} a method called Flow++ further improved the generative flows using variational dequantization and architecture design. It is a non-autoregressive model for unconditional density estimation. Hoogeboom~\etal~\cite{hoogeboom2019emerging} proposed emerging convolutions by chaining specific autoregressive convolutions which were invertible with receptive fields identical to standard convolution.  
Yuan and Kitani~\cite{yuan2020dlow} recently proposed DLow (Diversifying Latent Flows) for diverse human motion prediction from a pretrained deep generative model. It used learnable latent mapping functions to generate a set of correlated samples. Kumar~\etal~\cite{Kumar2019VideoFlowAF} replaced the standard unconditional prior distribution and introduced a latent dynamical system to predict future values of the flow model’s latent state. They used their model, VideoFlow, in the stochastic video prediction. 

Conditional normalizing flows based on RNN architecture have also been used for density estimation~\cite{kingma2016improved,oliva2018transformation}. Kingma~\etal~\cite{kingma2016improved}, for example, used RNNs to share parameters across the conditional distributions of autoregressive models which were non-linear generalizations of multiplication by a triangular matrix. Rasul~\etal~\cite{rasul2020multivariate} combined conditional normalizing flows with an autoregressive model, such as an RNN or an attention module. 

Masked autoregressive flows have also been proposed to share parameters without sequential computation of RNNs~\cite{papamakarios2017masked,ma2019macow}. MaCow~\cite{ma2019macow}, for example, restricted the local dependencies in a small masked kernel using masked convolutions with rotational ordering. It established a bijective mapping between input and output variables in a generative normalizing flows architecture. 
As discussed in~\cite{papamakarios2021normalizing}, masked autoregressive flows are universal approximators which can efficiently be evaluated. In particular, they can represent any autoregressive transformation and thus transform between any two distributions. These methods are attractive due to their simplicity and analytical tractability~\cite{papamakarios2021normalizing}, although their expressivity can be limited. 

In this work, we propose the use of masking with arbitrary orderings in convolutional networks for conditional normalizing flows. In convolutional networks, masking can be performed by multiplying the filter with a same-sized binary matrix, resulting in a kind of convolution often known as autoregressive or causal convolutions~\cite{van2016pixel,jain2020locally}.
We also propose factorizing the latent space into the time-dependent latent variables.
In this way, we extend flow-based generative models into the setting of spatio-temporal structured output learning. We specifically propose a new generative model which is conditioned on spatio-temporal dependencies of the input elements for learning autoregressive high-dimensional structured outputs.

\section{Preliminary Remarks}
\label{sec:background}
In this section, we introduce two concepts that are essential towards our proposed solution, namely Normalizing Flows and Masked Convolutions. The first is used to model multi-modal data distributions, while we use Masked Convolutions to devise the spatial autoregressive dependencies among input variables.

\subsection{Normalizing Flows}
\label{sec:Normalizing_flow}

A normalizing flow is a transformation of a random variable from a simple probability distribution (\eg Normal or Gaussian) into a more complicated distribution through a sequence of differentiable invertible mappings~\cite{hoogeboom2019emerging,kobyzev2020normalizing}. From a mathematical point of view, this results in new distributions from an initial density by a chain of parameterized, invertible, and differentiable transformations. 

Let $z\!=\!f(y)$ be a bijective mapping between variables $y$ and $z$ with a known and tractable probability density function, such as $p_\theta(z)\!=\! \mathcal{N}(z;0, \textbf{I})$, with parameters $\theta$.  
Also, let $g\!=\!f^{-1}$ be an invertible function and  
$y\!=\!g(z)$. 
The following equation is then obtained by using a change of variables:
\begin{equation}
\begin{aligned}
    &p_\theta(y) = p_\theta(z)|\det \frac{\partial z}{\partial y}| , 
\end{aligned}
\end{equation}
where $p_\theta(y)$ is a more complicated distribution obtained by multiplying $p_\theta(z)$ by the absolute of the Jacobian determinant. The function $f$ can be learned, but its choice is limited since the Jacobian determinant and the inverse of $f$ must be computationally tractable. In the generative formulation, the function $g$ is considered as a generator which pushes forward the base distribution $p_\theta(z)$ to a more complex density~\cite{kobyzev2020normalizing}. The function $f$ can also be defined as the inverse of $g$, which can move (or ``\emph{flow}'') in the opposite direction from a complex distribution towards a regular ``\emph{normalizing}'' form of the base density $p_\theta(z)$. This implies that an arbitrary complicated non-linear invertible (bijective) function $g$ can be used to generate any distribution $p_\theta(y)$ from any base distribution $p_\theta(z)$. This is called \emph{normalizing flows} since $f$ normalizes the data distributions. If $g\!=\!g_M\circ g_{M-1}\circ ... \circ g_1$ is defined as a composition of $M$ bijective functions, then 
$g$ is also
bijective, and the normalizing flows can be written as $f\!\!=\!\!f_1\circ f_2\circ ... \circ f_M$, which can transform the output variable $y$ to a latent variable $z$ drawn from a simpler distribution. 

In a particular neural network where the layers produce the intermediate representations, the log-likelihood of $y$ is computed using the following equation:
\begin{equation}
    \log p_\theta(y) = \log p_\theta(z)+\sum_{j=1}^M\log |\det(\frac{\partial r_j}{\partial r_{j-1}})| ,
\end{equation}
where $\{r_j\}_{j=1}^M$ denote $M$ intermediate representations, $r_0\!=\!y$ and $r_M\!=\!z$. 

Normalizing flows have been shown to be desirable solutions to address one-dimensional regression problems~\cite{kobyzev2020normalizing}. 
Here we extend the normalizing flows' abilities to spatio-temporal structured prediction, where the outputs are time dependent high-dimensional structured tensors.

\subsection{Masked Convolutions}
\begin{figure}[!htb]
    \centering
    \includegraphics[width=1\linewidth]{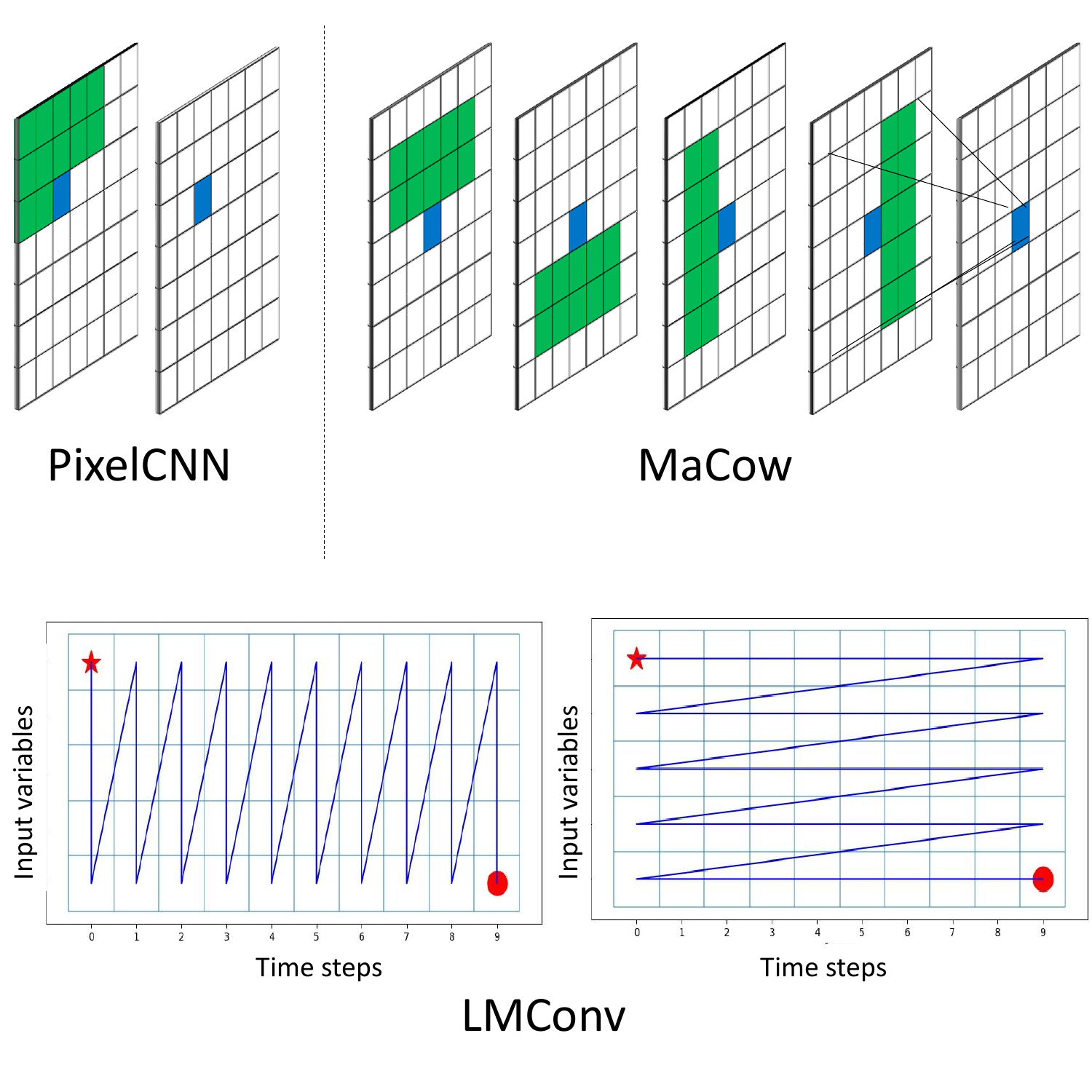}
    \caption{(Top) Receptive field with different mask orderings in 
    PixelCNN~\cite{van2016pixel,oord2016conditional}, and Macow~\cite{ma2019macow}. PixelCNN  masks the convolutions to generate each pixel by conditioning on the previously generated pixels in a raster scan ordering. Macow employs a rotational ordering by stacking multiple layers of convolutional flows. (Bottom) Orderings in time-dependent variables using LMConv~\cite{jain2020locally} which enables the arbitrary orders, allowing masks to be applied to the weights at each customized location. $\ast$ and $\bullet$ show the starting point and the stopping point, respectively. 
    }
    \label{fig:mask_1}
\end{figure}

Masking is a form of attention or autoregression,
in which the input is multiplied by a binary mask to either
emphasize or de-emphasize the relationship between certain 
elements, as well as imposing a specific ordering on the data.
Learning the
autoregressive structure of the observed data
is necessary to
model sequential dynamics.
For example, the previous time step will have an impact on the
current time step, which is an autoregressive relationship in the temporal domain. 
Other such relationships can also exist within a single frame in the spatial domain.
Traditionally, recurrent neural networks (RNNs) are used as generic and expressive autoregressive models for univariate outputs~\cite{hochreiter1997long,bahdanau2014neural,cho2014learning,rasul2020multivariate}. Convolutional Neural Networks (CNN) can also be used for autoregressive modeling. This is introduced in PixelCNN~\cite{van2016pixel,oord2016conditional} which outputs a conditional distribution at each image location. This is achieved by masking the convolutions to include only the valid context as defined by a specific set of pixels. In particular, each pixel is generated by conditioning on the previously generated pixels. 
The pixel generation is based on a specific fixed raster scan ordering, with pixels generated  row by row, and pixel by pixel within each row. 
MaCow~\cite{ma2019macow} used masks in a fixed
rotational ordering and stacked multiple layers of convolutional flows to capture a large receptive field. Figure~\ref{fig:mask_1} illustrates different mask orderings in PixelCNN and MaCow. 

Masked convolutions can be used to deal with the autoregressive nature of the inputs. Using the particular fixed orderings of PixelCNN or MaCow, however, can limit expressive dynamics. This has been alleviated by the recently proposed locally masked convolutions (LMConv)~\cite{jain2020locally}, which allow arbitrary orderings which can be tailored to a particular problem. In particular, we condition the output variables on the spatio-temporal features of the dynamic inputs by 
tailoring the LMConv masking strategy for our method, as illustrated in Figure~\ref{fig:mask_1}.

\section{Proposed Method}
\label{sec:proposed}

\subsection{Problem Formulation}

Let a multidimensional trajectory $q_{1:T} \!=\! (q^{(1)}, \dots, q^{(T)})$ be defined for a system consisting of $N$ interdependent and interacting entities, where $q^{(t)}= \{q^{(t)}_1, \dots, q^{(t)}_N\}$ denotes the set of features of all $N$ entities at time $t$, and  $q^{(t)}_i \in \mathbb{R}^D$ represents the $D$-dimensional feature vector of the $i$-th entity at time $t$.  
In one such example, trajectory data is $D=4$ dimensional,
with each $q^{(t)}_i$ comprising a 2D position and a 2D velocity component at time $t$ for the $i$-th entity.
Given the input sequence $x\!=\!\{q^{(1)},\dots,q^{(U)}\}$ with the observed history of $U$ time frames, the goal is to predict the sequence $y\!=\!\{q^{(U+1)},\dots, q^{(U+V)}\}$, where $T\!\!=\!\!U\!+\!V$. We thus aim to model a mapping from input frames at time steps $t\!=\!1,\dots ,U$ to a structured output consisting of the frames $t\!=\!(U\!+\!1), \ldots, (U\!+\!V)$. 

Interactions between a system's entities are investigated across time in order to model the system's dynamics. It is however common to study the trajectory prediction task instead, since there is no ground truth for the interactions between entities. Thus, to model the dynamics of a system, the future trajectory of its entities are predicted given the past trajectories~\cite{banijamali2021neural,graber2020dynamic,wang2021rdi,gong2021memory}.

We use conditional normalizing flows (CNFs)~\cite{winkler2019learning,trippe2018conditional} to model the conditional distribution $p(y|x)$ in a supervised manner,  where $x$ and $y$ respectively denote the input and output sequences in our model. We thus aim to predict the expected value of a label $y$ conditioned on observing associated features $x$.

\subsection{Our Method: MotionFlow}

An interesting intuition that informs our work is that the variables in the input and output sequences are correlated in a spatio-temporal space. 
We therefore propose to enhance the conditionals in CNFs to include time-dependent spatial features. Our method can be combined with any normalizing flows. In particular, we use flows with the steps including three layers of  \emph{actnorm} (which stands for activation normalization), \emph{invertible $1\!\times\! 1$ convolution}, and \emph{affine coupling}. 
In the Glow model, these layers are used with additional squeeze and split layers. 
Nevertheless, depending on the number of flow levels, it needs inputs and outputs with dimensions that are divisible by a factor of 2. In our method, we develop a more adaptable model by removing these layers. 

We utilize flow layers with additional conditioning layers to learn the relationships between spatio-temporal input features and structured output variables. The conditionals provide autoregressive priors on the weights and biases mapping from the hidden layers to the parameters of the normalizing flows.  
Masked convolutions with spatial and temporal orderings generate conditional weights for the steps of the normalizing flows. Notably, we process all $U$ input time steps in parallel in a single stage, which is in contrast to the RNN-based methods which require $U$ sequential stages to process all time steps. 

Although the prescribed model can capture the autoregressive characteristics and the spatial dependencies of the input sequence, the temporal dynamics of the output sequence are ignored if it is represented as a single datapoint. Inspired by VideoFlow~\cite{Kumar2019VideoFlowAF} which is a generative model for videos, we propose to learn a dynamic latent space where each time-dependent output variable depends on all previous latent states. We specifically propose temporally conditional prior distribution for structured output prediction. 

\begin{figure}[!t]
    \centering
    \includegraphics[width=1\linewidth]{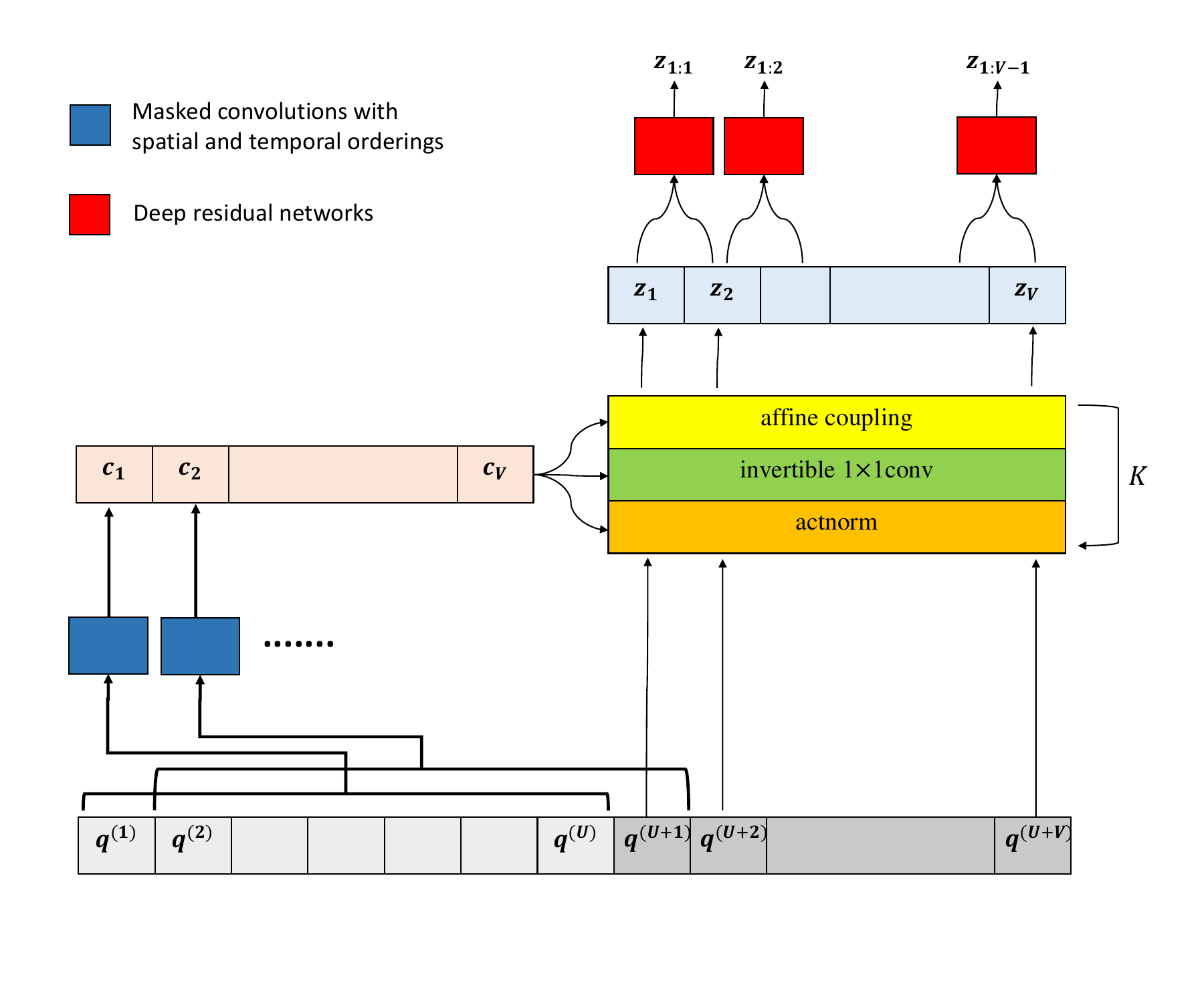}
    \caption{An overview of the MotionFlow (in the training time). Spatio-temporal input features are modeled using masked convolutions with spatial and temporal orderings. The correlations between time-dependent output variables are learned in the dynamic latent space, where each latent variable is temporally conditioned on its previous variables.}
    \label{fig:model}
\end{figure}

The overall pipeline is depicted in Figure~\ref{fig:model}. Given the input sequence, the output sequence is modeled in the latent space through normalizing flow layers. These layers are conditioned on the spatio-temporal features of the input sequence, which are extracted by the masked convolutions with different orderings. The latent space is also factorized to frame-wise latent variables, which their interconnections are modeled by deep residual networks.

In the following subsections, we discuss the main components of our method. An overview of the proposed architecture of our MotionFlow is represented in Figure~\ref{fig:model_detail}.

\subsection{Spatio-temporal Conditioner}

In CNFs, it is shown that by placing priors on the model distribution, the amount of expansion and contraction of the base distribution can be controlled in each step of the normalizing flows~\cite{trippe2018conditional,lu2020structured,wehenkel2021graphical}. 
As long as the steps of normalizing flows $f$ create a bijective map, they can be defined in any form. In general, $f$ can be conditioned on $\phi$ as follows:
\begin{equation}
f_\phi \!=\!f_{1_{\phi^1}}\circ f_{2_{\phi^2}}\circ  \dots f_{M_{\phi^M}}
\end{equation}
where each $\phi^i$ denotes a conditioner which constrains the Jacobian structure of the $f_i$. Thus, each $f_i$ which is partially parameterized by its conditioner must be invertible with respect to its input variable. It can be parameterized using a variety of approaches, such as~\cite{huang2018neural,durkan2019neural, trippe2018conditional} and~\cite{wehenkel2021graphical}. These methods however are used for one-dimensional regression problems. Recently, Lu and Huang~\cite{lu2020structured} used Glow~\cite{kingma2018glow} with extra neural networks to capture the relationship between input features and structured output variables. We further extend the CNFs for the structured spatio-temporal data, where $\phi^i$ are autoregressive conditioners. 

In our model, the conditioners include masked convolutions for masking at the feature level on the feature maps. We use locally masked convolution (LMConv)~\cite{jain2020locally}
to generate masks and use them as kernel weights for convolutions. By using masks in different orderings and stacking multiple autoregressive layers, our model can capture large receptive fields, and explore correlations in both the spatial and temporal dimensions in the input sequence.  

\begin{figure}[!t]
    \centering
    \includegraphics[width=1\linewidth]{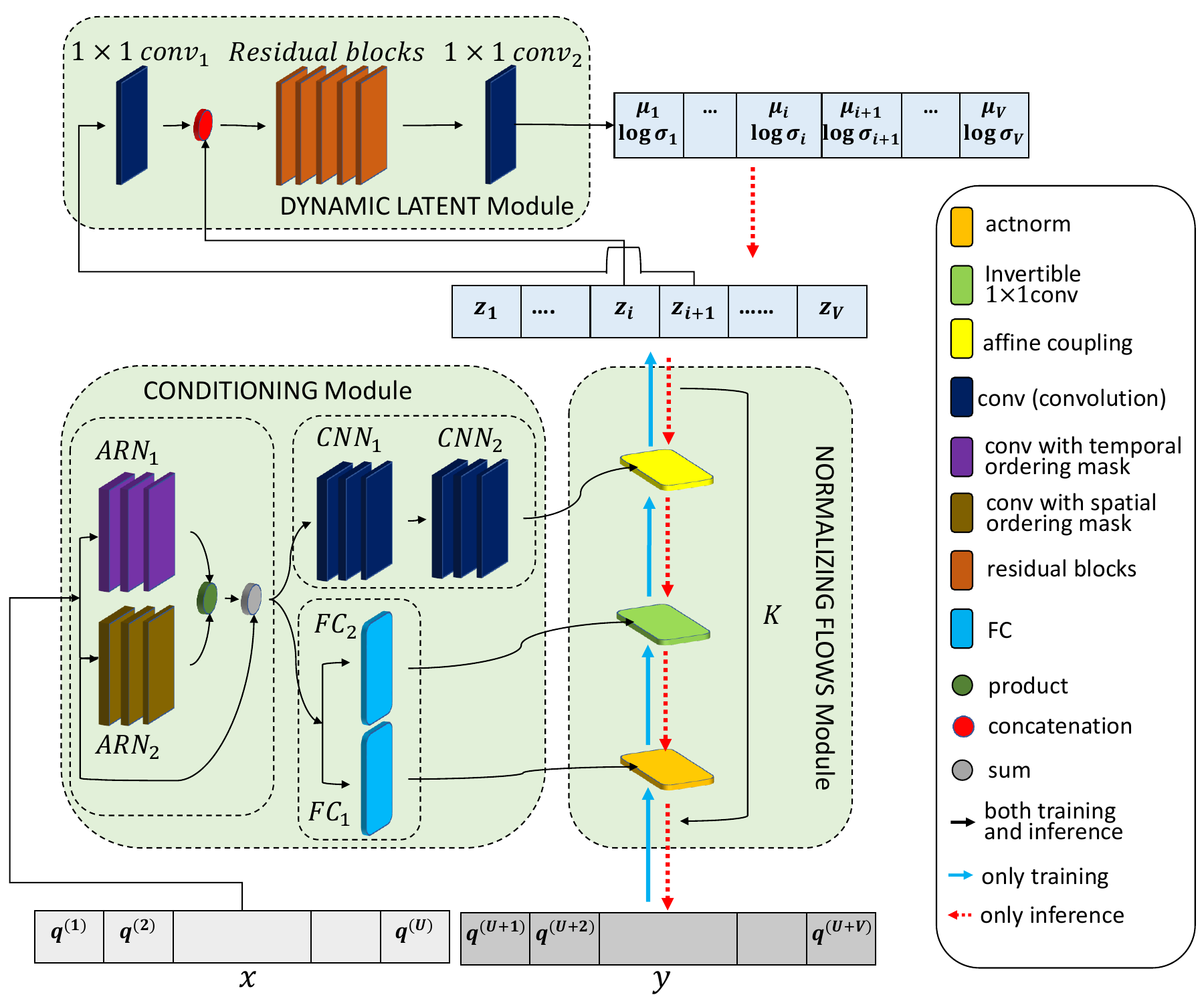}
    \caption{MotionFlow architecture. $ARN_1$ and $ARN_2$ denote the first and second autoregressive networks with convolutions utilizing spatial and temporal masks. The conditioning module including $ARN_1$, $ARN_2$, $CNN_1$, $CNN_2$, $FC_1$, and $FC_2$ provide weights and biases for the layers of normalizing flows. The latent space is factorized into time-dependent latent variables $z_i$ which their dynamic correlations are extracted using a deep residual network.}
    \label{fig:model_detail}
\end{figure}

As shown in Figure~\ref{fig:model_detail}, we use two autoregressive networks (\ie, $ARN_1$ and $ARN_2$) to learn the spatio-temporal dependencies in the input sequence. We use the masks with spatial and temporal orderings in these networks (see Figure~\ref{fig:mask_1}). Each autoregressive network comprises three convolution operations. One convolution with an initial mask operates on the input, and the other two convolutions use undilated and dilated kernel masks, respectively. The kernel size for all three convolutions is $3\times 3$. The only difference between the two autoregressive networks $ARN_1$ and $ARN_2$ being that their masks correspond to different orderings. 

The outputs of the two autoregressive networks are combined and fed to two fully connected layers and another convolution network. We apply fully connected networks to map the representations obtained by the conditioning submodule to determine the scale and bias parameters to initialize the conditioning for each step of the normalizing flows network. We use $FC_1$ for the actnorm networks and $FC_2$ for the invertible $1\!\times\! 1$ convolutional networks. 
The networks $CNN_1$ and $CNN_2$ build the scale and bias parameters for the affine coupling layers in the normalizing flows. These networks include convolution operations without masking. 

These autoregressive priors encode autoregressive characteristics of the input sequence for each normalizing flow layer by varying means and variances over the weights and biases connecting into the output units. Let the input and output of each layer of the normalizing flows be respectively denoted as $\alpha$ and $\beta$. We define the shape of each layer by $[h\!\times\!w\!\times\!c]$. The spatial and channel dimensions are then respectively denoted as $[h,w]$ and $c$. 

In the \emph{actnorm} layer, scale and bias parameters of each channel are used to perform an affine transformation of the activations on motion sequences. This operation is similar to batch normalization~\cite{ioffe2015batch}, which stabilizes the training of the network consisting of multiple flow operations. It is defined as:
\begin{equation}
    \beta_{i,j}=s\odot \alpha_{i,j}+b.
\end{equation}

Here $(i, j)$ denotes the spatial indices into tensors $\alpha$ and $\beta$, operator $\odot$ is the element-wise product, and $s$ and $b$ are two $1\!\times\!c$ vector parameters denoting the scale and the bias. In our method, these two vectors are generated from the output of $FC_1$, \ie, $s, b = FC_1(u)$, where $u$, the input to the fully connected network $FC_1$, is obtained by:
\begin{equation}
\begin{split}
    &c_1, c_2 = split(ARN_1 \odot ARN_2) \\
    &c_1 = c_1 - (mean(c_1) / std(c_1)) \\
    &c_3 = c_1 * sigmoid(c_2) \\
    &u = x + c_3, \\
\end{split}
\label{eq:pono}
\end{equation}
where $split()$ partitions a tensor into two equally sized chunks along the channel dimension. The second line in Eq. \ref{eq:pono} performs positional normalization~\cite{li2019positional} which is conducted based on positional statistics of mean and standard deviation to normalize the embeddings across the channel dimension. It enables masks to have a different number of ones at each spatial position.  

The actnorm layer is followed by an \emph{invertible 1 \!$\times$\! 1 convolutional} layer which generalizes the permutation operation on the motion sequences. It allows learnable reordering of channels in the input layer by incorporating a permutation along the channel dimension. It is defined as:
\begin{equation}
    \beta_{i,j}=W \alpha_{i,j},
\end{equation}
where $W$ denotes a $c\times c$ weight matrix. 
We initialize this matrix by using the output of the $FC_2$ as $W=FC_2(u)$, where $u$ is obtained similarly to the actnorm layer. Here, Eq. \ref{eq:pono} provides the input to the fully connected network $FC_2$.

An \emph{affine coupling} layer finally captures the correlations among spatial dimensions. It splits the input variables along the channel dimension into two halves, \ie, $\alpha_1$ and $\alpha_2$. It then adds a learned transformation of one half to the other half. 
More specifically, $u$ which is obtained via Eq. \ref{eq:pono} is used in the convolutional network $CNN_1$. The output of $CNN_1$ is concatenated to $\alpha_1$. This creates $h$ which is used in $CNN_2$ to generate $s$, and $b$ parameters for $\alpha_2$ to build $\beta_1$. The output of this layer is finally obtained by concatenating $\alpha_1$ and $\beta_1$. These operations are formulated as: 
\begin{equation}
\begin{aligned}
    & \alpha_1,\alpha_2=split(\alpha), \\
    & h = concat(\alpha_1, CNN_1(u)), \\
    & s,b = cross(CNN_2(h)), \\
    & \beta_1 = s\odot \alpha_2 + b,\\
    & \beta = concat(\alpha_1,\beta_1),
\end{aligned}
\label{eq:affine}
\end{equation}
where $split()$, $cross()$, and $concat()$ perform operations along the channel dimension. Similar to $split()$, $cross()$ partitions the tensor into two equally sized chunks, but the chunks are selected from even and odd indices separately. The size of the $s$ and $b$ vectors is the same as that of $\beta_1$.

\subsection{Dynamic Latent Space}
\label{sec:temporal}
We propose to model the temporal dependencies in the output sequence $y$ by imposing constraints on the prior $p_\phi(z)$. We specifically define frame-wise latent variables and leverage their interconnections in an autoregressive fashion. The  space of the latent variable $z$ is hence divided into disjoint latent subspaces per time frame: $z = \{z_t\}^V_{t=1}$, where $z_t$ denotes an invertible transformation of the corresponding frame $q^{(t)}$ of the output sequence $y$, \ie, $q^{(t)}=g_\phi(z_t)$. The latent space is then factorized as:
\begin{equation}
p_\phi(z) = \Pi_{t=1}^V p_\phi (z_t|z_{1:t-1}),
\end{equation}
where $z_{1:t-1}\!=\!\{z_{1},\dots, z_{t-1}\}$ denotes the latent variables of all time steps prior to the $t$-th step. We let each $p_\phi(z_t |z_{1:t-1})$ be a conditionally factorized Gaussian density:
\begin{equation}
p_\phi(z_t |z_{1:t-1})=\mathcal{N}(z_t;\mu_t,\sigma_t) 
\end{equation}
where $NN_\phi(z_{1:t-1})\!=\!(\mu_t, \log \sigma_t)$, is a deep residual network~\cite{he2016deep,Kumar2019VideoFlowAF}, which maps $z_{<t}$ to $(\mu_t, \log \sigma_t)$.   

Let $h_t$ be the tensor representing $z_t$ after the  affine coupling layer. We apply a $1 \times 1$ convolution over $h_t$ and concatenate this across channels to each latent variable from the previous timestep as:
\begin{equation}
\begin{aligned}
    &h_t = conv_1(h_t) \\
    &u_t = concat(h_{t-1}, h_t) \\
\end{aligned}
\end{equation}

We then transform it into $(\mu_t , \log \sigma_t)$ via a stack of residual blocks. Each 3D residual block includes three layers. The kernel
size for the first and third layers is $(2\times 3 \times 3)$. The second layer uses a Gated Activation Unit~\cite{van2016conditional} with two $1 \times 1 \times 1$ convolutions. We use a Gaussian distribution to initialize the first and second layer, while zero initialization is applied to the third layer. 
This is mainly performed for a stable optimization. 
Each residual block is duplicated with different dilation rates from $\{1,2,4\}$. The input and output of every block are then added together. We finally apply another $1 \times 1$ convolution on the output of the residual blocks sequence and achieve $(\Delta z_t , \log \sigma_t)$ as:
\begin{equation}
\begin{aligned}
    &u_t = res(u_t) \\
    &\Delta z_t , \log \sigma_t = conv_2(u_t) \\
\end{aligned}
\end{equation}
where $res()$ denotes the sequence of residual blocks. Eventually, $\Delta z_t$ is  added to $h_{t-1}$ (the tensor represnting $z_{t-1}$) to generate $\mu_t$ as:
\begin{equation}
    \mu_t=\Delta z_t + h_{t-1}. 
\end{equation}
The network can therefore learn the changes in the latent variables for each time step. 

As illustrated in Figure~\ref{fig:model_detail}, the latent variables $\{z_i\}_{i=1}^V$ and parameters $\{\mu_i,\log \sigma_i\}_{i=1}^V$ are created during the training stage. 
Blue and red arrows indicate the control flow for the training and inference, respectively. Black arrows indicate both training and inference.
The model is trained using the negative log-likelihood loss function as:
\begin{equation}
    \sum_{i=1}^V (\log p_\theta(z_i) + \log p_\phi(z_i) + \sum_{j=1}^M\log |\det(\frac{\partial r_j}{\partial r_{j-1}})|),
\end{equation}
where $r_M\!=\!z_i$. Notably, $\theta$ denotes the parameters learned by the normalizing flow module, and $\phi=(\mu_i,\log \sigma_i)$ are parameters learned by $NN_\phi(z_{1:i})$.  

We can use gradient-based optimization techniques such as gradient descent to optimize $y$. 
Once trained, our proposed model is capable of learning high-dimensional outputs with temporal and spatial interactions to predict future dynamics.

\subsection{Inference}
At inference time, after the model parameters are learned, output $y$ is predicted from an input $x$ by following the generative process of the model:
\begin{equation}
    y^* = \argmax_y p_\theta(y|x,z^*).
\end{equation}

We autoregressively generate the sequence $y=\{q^{(U+1)},\dots , q^{(U+V)}\}$ given the input sequence $x$. 
We initiate $(\mu_1, \log \sigma_1)$ by virtually using $q^{(U)}$ as one output time step. The generated $(\mu_1, \log \sigma_1)$ is then used to obtain $z_1$ and consequently frame $q^{(U+1)}$. In particular, the latent variable $z_1$ is modeled from $p_\theta(z_1)$ and propagated to obtain the corresponding output frame $q^{(U+1)}$. 
The input sequence is updated by the new predicted frame, and the procedure continues to obtain all output frames.

\section{Experiments}
\label{sec:experiments}
We evaluate the quantitative and qualitative aspects of the proposed method on both synthetic and realistic data on different tasks, including trajectory prediction, motion prediction, times series forecasting, and binary segmentation using structured prediction. 

We experiment with a simulated data and the NBA (National Basketball Association) dataset~\footnote{\url{https://github.com/linouk23/NBA-Player-Movements}} on the trajectory forecasting task in interacting systems. We then study the ability of our MotionFlow in motion prediction task on the CMU motion capture (CMU Mocap)~\footnote{\url{http://mocap.cs.cmu.edu/}} dataset~\cite{de2009guide}. As another task, we test our method on time series forecasting on 6 different datasets. Finally, we evaluate our MotionFlow on the binary segmentation task using structured prediction on the horse image dataset~\cite{borenstein2002class}.
We have found that MotionFlow achieves state-of-the-art performance by
leveraging the 
stochastic predictions of normalizing flow
with the deterministic constraints of the
latent space conditionals.
We also present ablation experiments on the key components of our model to quantify their impact.

\subsection{Implementation Details}

All models are trained using Adam optimizer~\cite{kingma2014adam} with learning rate 0.0001, weight decay 0.0005, $\beta_1=0.9$, and $\beta_2=0.999$. For all qualitative experiments and quantitative comparisons with the baseline methods, we use a batch size of 32, and train the network with early stopping condition on a single NVIDIA TITAN RTX GPU. We implement our models using PyTorch and make the code publicly available~\footnote{https://github.com/MohsenZand/MotionFlow} to reproduce the results. 

We have listed the network parameters and the details of our MotionFlow in Table~\ref{tab:convs}. The term $y_{ch}$ denotes the first dimension of the output, which varies depending on the experiment. In the motion prediction task, for example, it equals 24. Also, we set $K=8$ for all experiments.

\begin{table*}
\centering
\caption{Network parameters of the MotionFlow architecture. }
\begin{tabular}{cccccc}
\hline
&Type & Kernel & Channel & Masked & Kernel type \\
\hline
\hline
 \multirow{3}{*}{$ARN_1$ and $ARN_2$} & Conv 2d & $3\times 3 $ & 32 & \checkmark & initial \\
& Conv 2d & $3\times 3$ & 16 & \checkmark & undilated\\
& Conv 2d & $3\times 3$ & 4 & \checkmark & dilated\\
\hline
\multirow{3}{*}{$FC_1$}  & Linear & - & 64 & - & - \\
 & Linear & - & 64 & - & - \\
& Linear & - & $(2\times y_{ch})$ & - & - \\
\hline
\multirow{3}{*}{$FC_2$}  & Linear & - & 64 & - & - \\
 & Linear & - & 64 & - & - \\
& Linear & - & $(y_{ch})^2$ & - & - \\
\hline
\multirow{3}{*}{$CNN_1$} & Conv 2d & $3\times 3$ & 8 & - & - \\
& Conv 2d & $3\times 3$ & $y_{ch}/2$ & - & -\\
& Conv 2d & $3\times 3$ & $y_{ch}/4$ & - & -\\
\hline
\multirow{3}{*}{$CNN_2$} & Conv 2d & $3\times 3$ & 128 & - & - \\
& Conv 2d & $1\times 1$ & 128 & - & -\\
& Conv 2d & $3\times 3$ & $y_{ch}$ & - & -\\
\hline
$1\times 1 conv_1$ & Conv 2d & $1\times 1$ & $y_{ch}$ & - & -\\
\hline
\multirow{3}{*}{Residual block} & Conv 2d & $3\times 3$ & 512 & - & - \\
& Conv 2d & $1\times 1$ & 512 & - & -\\
& Conv 2d & $3\times 3$ & $(2 \times y_{ch})$ & - & -\\
\hline
$1\times 1 conv_2$ & Conv 2d & $1\times 1$ & $(2\times y_{ch})$ & - & -\\
\hline
\end{tabular}
\label{tab:convs}
\end{table*}

\begin{table*}[]
    \centering
    \caption{Mean squared error (MSE) in predicting future trajectories for simulations with 3 and 5 interacting objects.}
    \begin{tabular}{l|ccc|ccc}
    \hline
        & \multicolumn{3}{c|}{prediction steps for 3 particles} & \multicolumn{3}{c}{prediction steps for 5 particles}  \\
        Method & 1 & 15 & 25 & 1 & 15 & 25 \\
        \hline \hline
        \rowcolor[gray]{.9} 
        NRI~\cite{kipf2018neural} &  1.99e-5 & 1.50e-3 & 3.93e-3 & 2.67e-5 & 8.30e-3 & 2.05e-2 \\
        
        DNRI~\cite{graber2020dynamic} &  1.77e-5 & 3.48e-4 & 5.36e-4 & 2.10e-5 & 7.54e-4 & 1.80e-3 \\
        
        \rowcolor[gray]{.9} 
        MemDNRI~\cite{gong2021memory} &  1.26e-5 & 2.98e-4 & 4.56e-4 & - & - & - \\
        
        NRI-NSI~\cite{banijamali2021neural} &  \textbf{1.15e-5} & 3.20e-4 & 4.56e-4 & - & - & - \\
        
        \rowcolor[gray]{.9} 
        Ours (MotionFlow) & 1.55e-5 & \textbf{2.88e-4} & \textbf{4.32e-4} & \textbf{2.05e-5} & \textbf{7.35e-4} & \textbf{1.28e-3} \\
        
        \hline
    \end{tabular}
    \label{tab:synthetic}
\end{table*}

\subsection{Trajectory Prediction}
To evaluate the effectiveness of our method in dealing with dynamic and evolving interacting systems with multiple heterogeneous, interactive agents, we test it on both controlled synthetic and real world trajectory data. 

\subsubsection{Synthetic Simulations}
We evaluate our model on a synthetic dataset consisting of moving particles with dynamic relations. Following~\cite{kipf2018neural,gong2021memory,graber2020dynamic}, we construct a system with trajectories of three particles. 
Two particles move with a fixed velocity, while the third is given a random velocity. It is pushed away by the other two particles whenever the distance between them is less than one. Furthermore, the interactions between the particles change dynamically based on their distances. 

Results are reported in Table~\ref{tab:synthetic}. We compare the trajectory prediction mean squared error (MSE) with NRI (neural relational inference)~\cite{kipf2018neural}, DNRI (dynamic neural relational inference)~\cite{graber2020dynamic}, MemDNRI (memory-augmented dynamic neural relational inference)~\cite{gong2021memory}, and NRI-NSI (neural relational inference with node-specific information)~\cite{banijamali2021neural}. All models are provided with the first 10 frames as inputs, and the remaining 25 frames are predicted. The results show that MotionFlow achieves the best performance,
with the lowest MSE,
at the intermediate 15 and the final 25 steps.
While the other methods try to identify the relations between interacting objects in an unsupervised manner, in contrast our method uses a supervised model for predicting dynamic relations through trajectory prediction without explicitly estimating the relations. NRI, for example, treats interactions as latent variables in a variational auto-encoder (VAE)~\cite{welling2013auto} and uses inferred relations types on a graph to predict the future trajectories. It is however limited to static relations across the observed trajectory. Other methods use different strategies to leverage the time-dependent relation, such as using LSTM in DNRI. Our method however conditionally generates trajectories by assessing the input as a whole and updates the input at each time step. 

\begin{figure}[!t]
    \centering
    \includegraphics[width=1\linewidth]{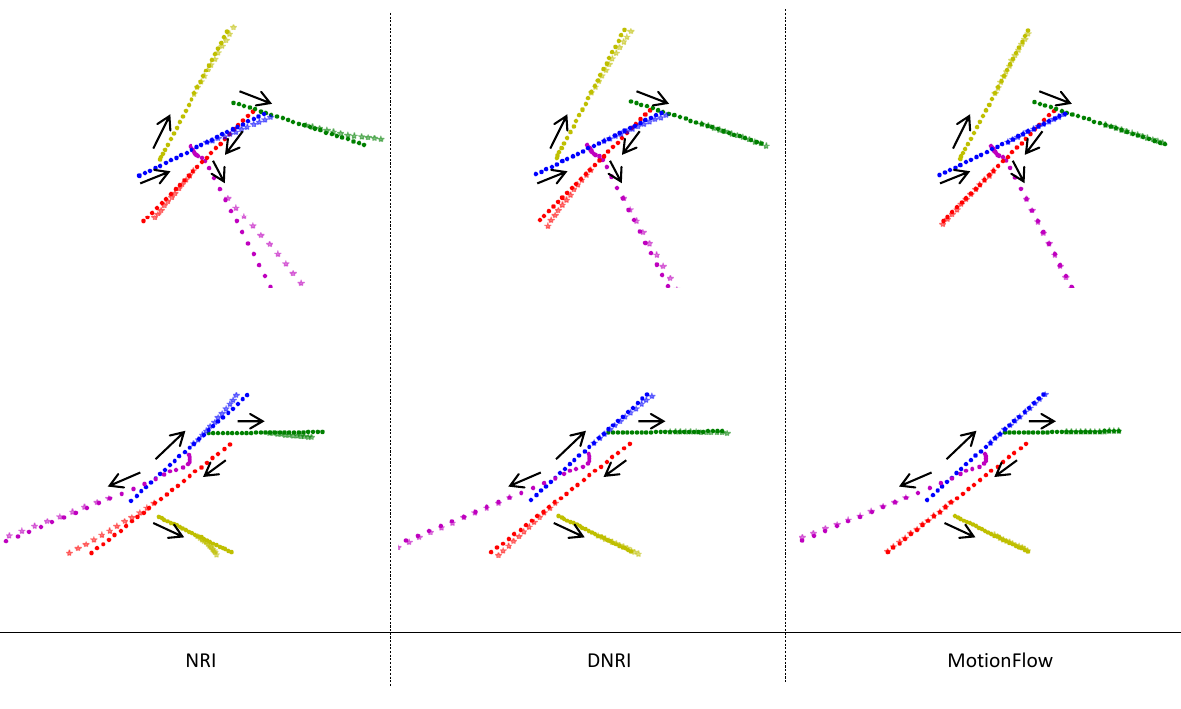}
    \caption{Visualization of the dynamic trajectories on the simulation data. The solid circles denote the ground truth, and the semi-transparent asterisk symbol indicate the predicted locations. Three particles with red, blue, and green colors move with a fixed velocity, while the other two particles in yellow and purple are given a random velocity and pushed away by the other particles when they are close to each other. The movement direction is shown by black arrows.}
    \label{fig:sim}
\end{figure}

To further investigate the capacity of our method, we test it on another synthetic dataset with 5 particles. 
The velocity of the three particles a, b, and c is fixed, but the other two particles, d and e move with a random velocity. 
When the distance between a and b is less than one, they push the particle d away. All the particles a, b, and c have an impact on particle e. These particles push e away whenever their corresponding distance is less than one. 
We evaluate our method against NRI~\cite{kipf2018neural} and DNRI~\cite{graber2020dynamic} on this dataset, and report the results in Table~\ref{tab:synthetic}. 
In Figure~\ref{fig:sim}, we illustrate one example of the prediction results, where our method is able to better estimate the trajectories of all five particles, compared to NRI and DNRI.

\subsubsection{NBA dataset}
The NBA dataset uses actual player tracking information to represent the position trajectories of 10 interacting basketball players and the ball. We follow the same protocol as~\cite{sun2022interaction} and use 300k multi-agent trajectories.
Given a history of 24 time steps (10s), the challenge is to forecast 10 future time steps (4s). 

We evaluate our method in terms of the widely used average displacement error (ADE) and final displacement error (FDE)~\cite{kothari2021human,sun2022interaction}. ADE measures the average mean square error (MSE) over all time steps between the ground truth trajectory and the predicted trajectory. FDE is defined as the distance between the predicted final position and the final ground truth position at the last time step. 

The prediction performance on future trajectories are averaged over three independent runs and reported in meters. 
All baseline results are taken from~\cite{sun2022interaction}. 
As indicated in Table~\ref{tab:bball}, our method achieves 0.763 ADE and 1.509 FDE, which are the best and second-best performances among the baseline methods, respectively.
Our spatio-temporal modeling of the input and output sequences is primarily responsible for the performance gain, particularly for ADE. With MotionFlow,
it is possible to predict trajectories by capturing the autoregressive characteristics and the spatial dependencies of the input and output sequences.

\begin{figure}[!t]
    \centering
    \includegraphics[width=0.9\linewidth]{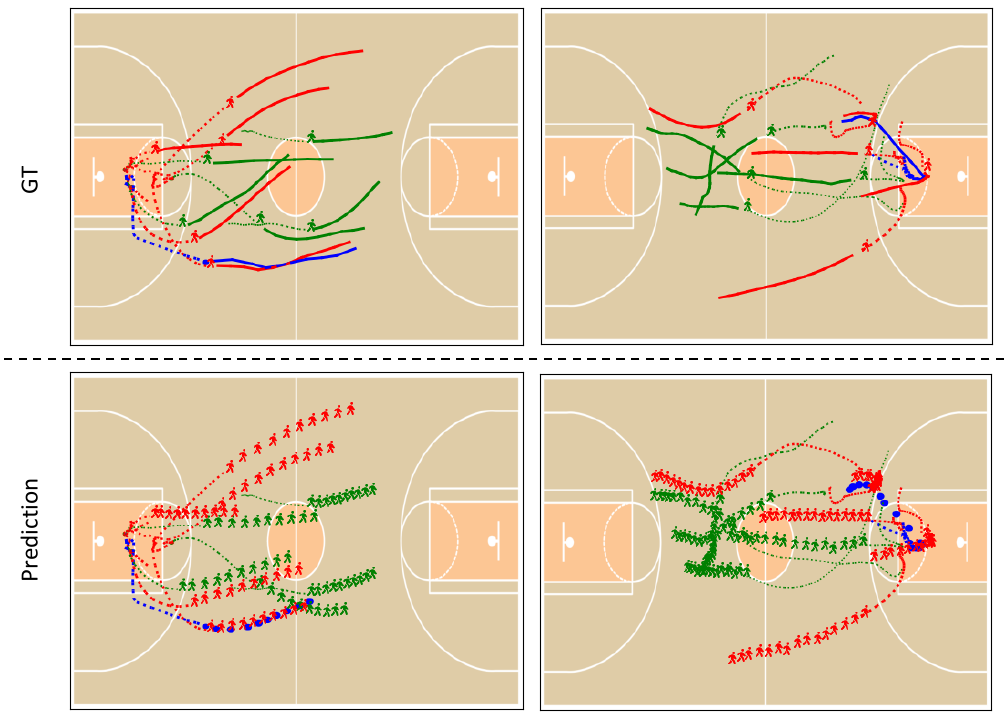}
    \caption{Visualization of predicted trajectories of two testing cases of NBA dataset. Historical trajectories are shown as dashed lines, while ground truths are shown as solid lines.
    Red and green are used to symbolize players on different teams, while blue is used to represent the ball.}
    \label{fig:bball_results}
\end{figure}

In Figure~\ref{fig:bball_results}, we illustrate predicted trajectories for a test case from NBA dataset. We can see that our method is able to predict smooth and realistic trajectories for all agents.

\begin{table*}[]
    \centering
    \caption{Trajectory prediction results in terms of ADE and FDE (mean ± std, over 3 runs) on the NBA dataset. The best performance is highlighted in bold}
    \begin{tabular}{lccccc}
        \hline
         & NRI~\cite{kipf2018neural} & EvolveGraph~\cite{li2020evolvegraph} & fNRI~\cite{webb2019factorised} & IMMA~\cite{sun2022interaction} & Ours (MotionFlow) \\
        \hline \hline 
        \rowcolor[gray]{.9} ADE $\downarrow$ & 0.946 $\pm$ 0.005 & 0.896 $\pm$ 0.009 & 0.804 $\pm$ 0.004 & \underline{0.769 $\pm$ 0.009} & \textbf{0.763 $\pm$ 0.005} \\
        FDE $\downarrow$ & 1.818 $\pm$ 0.017 & 1.695 $\pm$ 0.016 & 1.517 $\pm$ 0.012 & \textbf{1.438 $\pm$ 0.019} & \underline{1.509 $\pm$ 0.005} \\
         \hline
    \end{tabular}
    \label{tab:bball}
\end{table*}

\begin{table*}
\centering \scriptsize
\setlength{\tabcolsep}{2.5pt}
\caption{Prediction results over 8 action categories of the CMU Mocap dataset. The results are reported in terms of Mean Per Joint
Position Error (MPJPE) in millimeter. The best performance is highlighted in bold. Our method achieves the best performance in most time
horizons for all activities }
\begin{tabular}{l|ccccc|ccccc|ccccc|ccccc}
\hline
actions & \multicolumn{5}{|c|}{basketball} & \multicolumn{5}{|c|}{basketball signal} & \multicolumn{5}{|c|}{directing traffic} & \multicolumn{5}{|c}{jumping}
 \\
\hline
milliseconds & 80 & 160 & 320 & 
400 & 1000 & 80 & 160 & 320 & 
400 & 1000 & 80 & 160 & 320 & 
400 & 1000 & 80 & 160 & 320 & 
400 & 1000 \\
\hline
 
Seq2seq~\cite{martinez2017human}
& 15.45 & 26.88 & 43.51 & 
49.23 & \textbf{72.83} & 20.17 & 32.98 & 42.75 & 
44.65 & 60.57 & 20.52 & 40.58 & 75.38 & 
90.36 & 153.12 & 26.85 & 48.07 & 93.50 & 
108.90 & 162.84 \\

\rowcolor[gray]{.9} 
TrajDep~\cite{mao2019learning}
& 11.68 & 21.26 & 40.99 & 
50.78 & 97.99 & 03.33 & 06.25 & 13.58 & 
17.98 & 54.00 & 06.92 & 13.69 & 30.30 & 
39.97 & 114.16 & 17.18 & 32.37 & 60.12 & 
72.55 & 127.41 \\

MSR-GCN~\cite{dang2021msr}
& \textbf{10.28} & \underline{18.94} & \textbf{37.68} & 
\underline{47.03} & {86.96} & \underline{03.03} & \underline{05.68} & \underline{12.35} & 
\textbf{16.26} & \underline{47.91} & \underline{05.92} & \underline{12.09} & \underline{28.36} &
\underline{38.04} & \underline{111.04} & \underline{14.99} & \underline{28.66} & \textbf{55.86} & 
\underline{69.05} & \underline{124.79} \\

\rowcolor[gray]{.9} 
Ours (MotionFlow) 
& \underline{11.22} & \textbf{18.35} & \underline{38.57} & 
\textbf{45.64} & \underline{84.27} & \textbf{03.02} & \textbf{05.62} & \textbf{12.24} & 
\underline{16.48} & \textbf{47.01} & \textbf{05.88} & \textbf{12.05} & \textbf{27.92} & 
\textbf{37.76} & \textbf{108.25} & \textbf{14.52} & \textbf{28.37} & \underline{60.02} & 
\textbf{68.60} & \textbf{122.16} \\\hline\hline

actions & \multicolumn{5}{|c|}{running} & \multicolumn{5}{|c|}{soccer} & \multicolumn{5}{|c|}{walking} & \multicolumn{5}{|c}{washing window}
 \\
\hline

milliseconds & 80 & 160 & 320 & 
400 & 1000 & 80 & 160 & 320 & 
400 & 1000 & 80 & 160 & 320 & 
400 & 1000 & 80 & 160 & 320 & 
400 & 1000 \\
\hline

Seq2seq~\cite{martinez2017human}
& 25.76 & 48.91 & 88.19 & 
100.80 & 158.19 & 17.75 & 31.30 & 52.55 & 
61.40 & 107.37 & 44.35 & 76.66 & 126.83 & 
151.43 & 194.33 & 22.84 & 44.71 & 86.78 & 
104.68 & 202.73 \\

\rowcolor[gray]{.9} 
TrajDep~\cite{mao2019learning} 
& 14.53 & 24.20 & 37.44 & 
41.10 & 51.73 & 13.33 & 24.00 & 43.77 & 
53.20 & 108.26 & 06.62 & 10.74 & \underline{17.40} & 
\underline{20.35} & \underline{34.41} & 05.96 & 11.62 & \textbf{24.77} &
\underline{31.63} & \textbf{66.95} \\

MSR-GCN~\cite{dang2021msr}
& \underline{12.84} & \underline{20.42} & \underline{30.58} & 
\textbf{34.42} & \textbf{48.03} & \textbf{10.92} & \underline{19.50} & \underline{37.05} & 
\textbf{46.38} & \underline{99.32} & \textbf{06.31} & \underline{10.30} & 17.64 & 21.12 & 39.70 & \underline{05.49} & \underline{11.07} & 25.05 & 32.51 & 71.30 \\

\rowcolor[gray]{.9} 
Ours (MotionFlow) 
& \textbf{10.13} & \textbf{20.31} & \textbf{28.42} & 
\underline{38.40} & \underline{50.88} & \underline{12.09} & \textbf{18.98} & \textbf{36.22} & 
\underline{48.50} & \textbf{97.49} & \underline{06.52} & \textbf{10.15} & \textbf{16.28} & 
\textbf{20.05} & \textbf{34.11} & \textbf{05.46} & \textbf{11.01} & \underline{24.80} & 
\textbf{30.95} & \underline{69.62} \\\hline
\end{tabular}

\label{table:cmu}
\end{table*}

\subsection{CMU Mocap} 

CMU motion capture (CMU Mocap)~\cite{de2009guide} is a challenging dataset containing 8 action categories performed by 144 subjects. 
Each pose is represented by 38 body joints, among which we preserve 24 joints in our experiments. 

As opposed to existing methods which extract a fixed-length sequence of frames from each longer dataset sequence, we utilize a sliding window strategy over the sequences. We extract sequences with the length of 3 seconds (150 frames with FPS = 50). The stride is also selected as 10 frames. Our data generator then creates a 2-second window from a random frame in the 3-second sequence. The first second is used as the input sequence, and we predict the following 1-second sequence. 

\begin{figure*}[!t]
    \centering
    \includegraphics[width=0.9\linewidth]{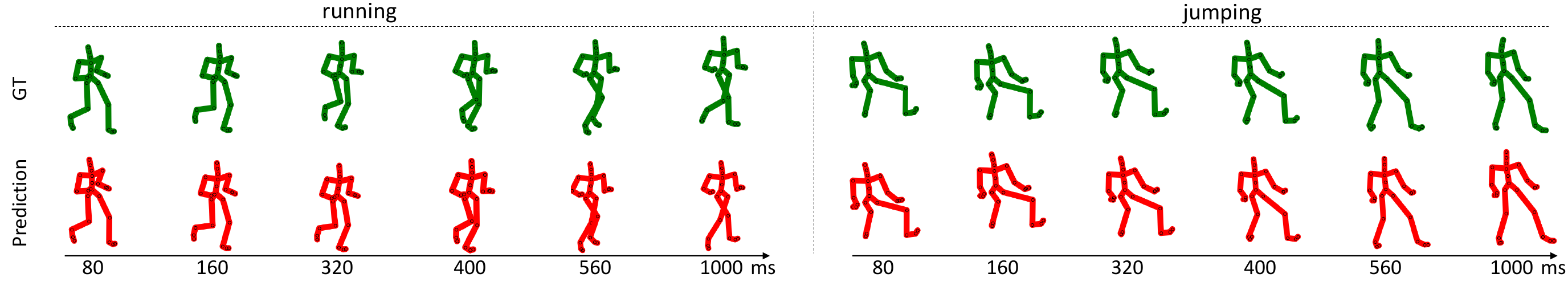}
    \caption{Qualitative analysis of short-term and long-term predictions on a sample 'basketball` sequence. The ground-truth and predictions are illustrated by green and red color, respectively. The results demonstrate that MotionFlow predicts both short-term and long-term high-quality motions. }
    \label{fig:motion_pred}
\end{figure*}

We use 
seq2seq~\cite{martinez2017human},
TrajDep (trajectory dependencies)~\cite{mao2019learning}, 
and MSR-GCN (multi-scale residual graph convolution networks)~\cite{dang2021msr} for comparison. The prediction results are shown in Table~\ref{table:cmu} in terms of Mean Per Joint Position Error (MPJPE) in millimeter which is the most widely used evaluation metric for motion prediction~\cite{dang2021msr,mao2019learning}. It can be seen that MotionFlow achieves the best performance in most time horizons. Other methods such as MSRGCN model the human motion deterministically, which are regressed to the mean pose and generate artifacts. The lack of motion diversity also leads to repetitive patterns in predicted motions.  
For a realistic motion prediction, however, the model should balance the deterministic and stochastic temporal representations. 
It should also be noted that these methods are specially designed for motion prediction and use human skeleton features (such as decomposing the human pose to a series of poses from fine to coarse scale, by grouping closer joints together and replacing the group with a single joint, which is performed by MSR-GCN) whereas our method generalizes the spatio-temporal prediction tasks. 
In Figure~\ref{fig:motion_pred}, we illustrate some results for both short-term and long-term motion prediction.

\begin{table}[]
    \centering
    \caption{Details of datasets used in time series forecasting experiment}
    \begin{tabular}{lcccc}
        \hline
        Dataset & Dimension & Total time steps & Prediction length \\
        \hline \hline 
        EXCHANGE & 8 & 6,071 & 30 \\
        \rowcolor[gray]{.9}SOLAR & 137 & 7,009 & 24 \\
        ELECTRICITY & 370 & 5,790 & 24 \\ 
        \rowcolor[gray]{.9}TRAFFIC & 963 & 10,413 & 24 \\
        TAXI & 1,214 & 1,488 & 24 \\ 
        \rowcolor[gray]{.9}WIKIPEDIA & 2,000 & 792 & 30 \\
         \hline
    \end{tabular}
    \label{tab:datasets}
\end{table}

\subsection{Time Series Forecasting}
For the time series forecasting task, we used 6 different datasets including Exchange~\cite{lai2018modeling}, Solar~\cite{lai2018modeling}, Electricity~\footnote{\url{https://archive.ics.uci.edu/ml/datasets/ElectricityLoadDiagrams20112014}}, Traffic~\footnote{\url{https://archive.ics.uci.edu/ml/datasets/PEMS-SF}}, Taxi~\footnote{\url{https://www1.nyc.gov/site/tlc/about/tlc-trip-record-data.page}}, and Wikipedia~\footnote{\url{https://github.com/mbohlkeschneider/gluon-ts/tree/mv_release/datasets}}. We process all datasets in the same way as~\cite{salinas2019high, rasul2020multivariate}. The details of all datasets are shown in Table~\ref{tab:datasets}. 

Exchange dataset contains daily exchange rates between 8 currencies.
In Solar dataset, 137 stations in Alabama State are represented by their hourly photovoltaic output.
Electricity dataset includes 370 users' power consumption in hourly time series. 
There are 963 San Francisco automobile lanes that have an hourly occupancy rate between 0 and 1 in the Traffic dataset. 
Taxi dataset provides spatio-temporal traffic time series of New York taxi rides gathered at 1214 sites once every 30 minutes in January 2015 (training set) and January 2016 (test set). 
Daily page views of 2000 Wikipedia pages create the Wikipedia dataset. Following~\cite{salinas2019high} and \cite{rasul2020multivariate}, the data from each dataset is divided into a training and test set utilizing sliding windows for the test set and all data before a specific date for the training set.

\begin{table*}[]
    \centering
    \caption{Comparison results in terms of MSE (lower is better) on the test sets of different time series datasets}
    \begin{tabular}{l|cccccc}
        \hline
        & \multicolumn{6}{c}{Datasets} \\
        Method & Exchange & Solar & Electricity & Traffic & Taxi & Wikipedia \\
        \hline \hline
        
        \rowcolor[gray]{.9} Vec-LSTM ind-scaling & $\mathbf{1.6 \times 10^{-4}}$ & $9.3 \times 10^2$ & $2.1 \times 10^5$ & $6.3 \times 10^{-4}$ & $7.3 \times 10^1$ & $7.2 \times 10^7$ \\
        
        Vec-LSTM lowrank-Copula & $1.9 \times 10^{-4}$ & $2.9 \times 10^3$ & $5.5 \times 10^6$ & $1.5 \times 10^{-3}$ & $5.1 \times 10^1$ & $3.8 \times 10^7$ \\
        
        \rowcolor[gray]{.9} GP Copula & $1.7 \times 10^{-4}$ & $9.8 \times 10^2$ & $2.4 \times 10^5$ & $6.9 \times 10^{-4}$ & $3.1 \times 10^1$ & $4.0 \times 10^7$ \\
        
        LSTM Real-NVP & $2.4 \times 10^{-4}$  & $9.1 \times 10^2$ & $2.5 \times 10^5$ & $6.9 \times 10^{-4}$ & $2.6 \times 10^1$ & $4.7 \times 10^7$ \\
        
        \rowcolor[gray]{.9} LSTM MAF & $3.8 \times 10^{-4}$ & $9.8 \times 10^2$ & $1.8 \times 10^5$ & $\mathbf{4.9 \times 10^{-4}}$ & $2.4 \times 10^1$ & $3.8 \times 10^7$ \\
        
        Transformer MAF & $3.4 \times 10^{-4}$ & $9.3 \times 10^2$ & $2.0 \times 10^5$ & $5.0 \times 10^{-4}$ & $4.5 \times 10^1$ & $\mathbf{3.1 \times 10^7}$ \\

        \rowcolor[gray]{.9} Ours (MotionFlow) & $1.7 \times 10^{-4}$ & $\mathbf{9.0 \times 10^2}$ & $\mathbf{1.6 \times 10^5}$ & $\mathbf{4.9 \times 10^{-4}}$ & $\mathbf{2.3 \times 10^1}$ & $3.5 \times 10^7$\\
        
        \hline
    \end{tabular}
    \label{tab:time_s}
\end{table*}

We compare our method with the 
models from~\cite{salinas2019high} and \cite{rasul2020multivariate}. They include different architectures, distributions, and transformations with low-rank Gaussian Copula Process~\cite{salinas2019high} and autoregressive conditioned normalizing flows~\cite{rasul2020multivariate}. Vec-LSTM ind-scaling
is a single LSTM with independent distribution and mean scaling transformation. Vec-LSTM lowrank-Copula is based on the same LSTM architecture but low-rank normal distribution with copula transformation are used~\cite{salinas2019high}. 
In GP model, a Gaussian process model is parametrized by unrolling the LSTM on each dimension prior to the joint distribution reconstruction. GP Copula uses a GP model with copula transformation. In~\cite{rasul2020multivariate}, LSTM is used with two normalizing flows Real NVP or MAF (denoted as LSTM Real-NVP and LSTM MAF in Table~\ref{tab:time_s}), where multiple layers of a conditional flow module (Real NVP or MAF) are stacked together with the RNN. MAF is also used with a self-attention mechanism in Transformer MAF model. 

In Table~\ref{tab:time_s}, we show the comparison results in terms of MSE (mean squared error) over all time series dimensions and over the entire predicted sequence. 
Our model achieves the best or the second-best performance on all datasets. The accuracy of our model is mostly due to capturing dependencies between all variables over time. The flexibility of  normalizing flows, which can adapt to a wide range of high-dimensional data distributions and their robustness are beneficial in time series forecasting. This can be seen in~\cite{rasul2020multivariate}, the other normalizing flows-based method that has been applied to time series analysis. Our spatio-temporal feature representation and dynamic latent space are however more suited to handle dynamic interactions.

\begin{table}[]
    \centering
    \caption{Segmentation results on Weizmann horses test dataset. The input size for all models is $32 \times 32$ pixels. }
    \begin{tabular}{lc}
        \hline
        Method & 
        Mean IoU (\%) \\
        \hline \hline
        
        \rowcolor[gray]{.9} FCN (baseline) & 
        78.7 \\
        
        DVN~\cite{gygli2017deep} & 
        84.0  \\
        
        \rowcolor[gray]{.9} c-Glow~\cite{lu2020structured} & 
        81.2 \\
        
        ALEN~\cite{pan2020adversarial} & 
        85.4 \\
        
        \rowcolor[gray]{.9} Ours (MotionFlow) & 
        \textbf{89.2} \\ 
        
         \hline
    \end{tabular}
    \label{tab:seg}
\end{table}

\subsection{Binary Segmentation}
To characterize the generality of our method, we have experimented on a binary segmentation task, which includes an element of structured prediction,
without any time-dependencies.
We use the Weizmann Horse image dataset~\cite{borenstein2002class} which is a common dataset for binary image segmentation and structured prediction evaluation. It consists of 328 images of horses and their binary segmentation masks. Following the experimental protocols of~\cite{lu2020structured}, we split the dataset and report the results on the test set. We also resize the images and masks to be $32 \times 32$ pixels and duplicate masks three times to generate 3-channels deep. 
We compare our method with a baseline FCN (fully convolutional network), and structured prediction algorithms of DVN (deep value networks)~\cite{gygli2017deep}, c-Glow (conditional Glow)~\cite{lu2020structured}, and ALEN (adversarial localized energy network)~\cite{pan2020adversarial} in terms of IoU (intersection over union).

To avoid the local optima due to the multi-modality and non-convexity of the density distribution in the structured prediction task and to converge faster, we follow a sample-based approximate strategy for inference~\cite{lu2020structured}. A set of samples drawn from $p_Z(z)$ are defined as $\{z_1, ..., z_S\}$. The output can then be estimated by averaging over these variables:
\begin{equation}
    y^*\approx \frac{1}{S} \sum_{i=1}^S g_{x,\phi}(z_i),
\end{equation}
where we empirically set the number of samples to 10. 

\begin{figure}[!t]
    \centering
    \includegraphics[width=1\columnwidth]{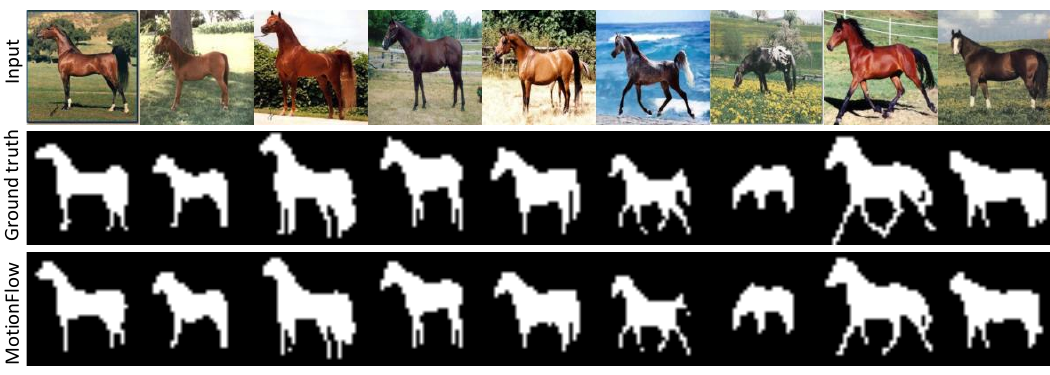}
    \caption{Segmentation results of MotionFlow on Weizmann horses test samples. }
    \label{fig:seg}
\end{figure}

As shown in Table~\ref{tab:seg}, our method achieves the highest IoU among the comparison methods with a large margin of 3.8\% to the second-best algorithm. c-Glow which is a conditional generative model based on the Glow architecture outperforms the feed-forward deep model of FCN. DVN and ALEN are deep energy-based models which use energy functions to capture the dependencies among output labels. We observe that ALEN with an adversarial learning framework outperforms DVN. Our method however shows stronger segmentation results since it generates more accurate conditional samples. Some qualitative results are shown in Figure~\ref{fig:seg}, where we observe the capability of MotionFlow in generating high-quality predictions. 
It satisfactorily works on the low resolution images of $32 \times 32$ pixels, where small details such as the legs are often obscured in the RGB images.

\begin{table*}
\centering
\caption{Ablation results over 'basketball` and 'walking` actions on CMU Mocap dataset. The
results are reported in terms of Mean Per Joint Position Error (MPJPE)
in millimeter. The best performance is highlighted in bold. A, B, and C
denote masked convolutions, dynamic latent space, and residual
networks, respectively. A letter is checked if its corresponding
component is used in our architecture.}
\begin{tabular}{ccc|ccccc|ccccc}
\hline
&& & \multicolumn{5}{c|}{Basketball} & \multicolumn{5}{c}{Walking}
 \\
\cline{4-13}
A & B & C  & 80 & 160 & 320 & 400 & 1000 & 80 & 160 & 320 & 400 & 1000 \\
\hline
\hline

\xmark & \cmark & \cmark
& 13.68 & 23.26 & 48.99 & 55.78 & 104.33 & 10.45 & 19.92 & 37.17 & 61.73 & 113.69 \\

\rowcolor[gray]{.9} \cmark & \xmark & \xmark & 15.45 & 26.88 & 48.51 & 57.23 & 110.17 & 12.26 & 25.83 & 38.45 & 65.63 & 121.58 \\

\cmark & \cmark & \xmark & 12.85 & 20.62 & 41.18 & 46.95 & 88.37 & 10.65 & 13.24 & 23. 31 & 36.70 & 72.69 \\

\rowcolor[gray]{.9} \cmark & \cmark & \cmark & \textbf{11.22} & \textbf{18.35} & \textbf{38.57} & \textbf{45.64} & \textbf{84.27} & \textbf{6.52} & \textbf{10.15} & \textbf{16.28} & \textbf{20.05} & \textbf{34.11} \\

\hline

\end{tabular}
\label{table:ablation}
\end{table*}

\subsection{Ablation Studies}
To evaluate the contribution of different aspects of the proposed method, we conduct ablation studies on the motion prediction task,
which is challenging as it includes both high dimensional
spatial and temporal features. 
We use the CMU Mocap dataset, and have selected \emph{'basketball`} and \emph{'walking`} as two representative actions in our experiment. 
In particular, we study the impact of (A) spatio-temporal conditioning using masked convolutions, (B) temporal skip connections in the latent space, and (C) residual network. 

\textbf{(A) Spatio-temporal conditioning} provides the scale and bias parameters to initialize the conditioning for each step of the normalizing flows network. We replace the masked convolutions with two convolutional layers, each consisting of a 2D convolution with kernel size $3 \times 3$ and a channel size of 256. The results are shown in the first row in Table~\ref{table:ablation}. Comparing the first and fourth rows of the table indicates that spatial-temporal conditioning has the second largest impact of any single component.

\textbf{(B) Temporal connections in the latent space} models the temporal dependencies in the output sequence in the latent space. We turn the prior factorization back to the standard prior $z$. As a result, (C) is inevitably not further required in this experiment. In Table~\ref{table:ablation}, this is depicted in the second row, where only A is checked.
Comparing the second and fourth rows of the table shows that temporal connections has the largest impact on performance of any single component. 
Furthermore, output sequences with more stochastic features are predicted by removing the deterministic constraints of the latent space conditionals.

\textbf{(C) A residual network } is utilized to extract the interdependencies between time-dependent latent variables. As shown in Table~\ref{table:ablation}, the impact of this network is much lower than the other two parts. 

In another experiment, we evaluate our network initialization strategy explained in Sec. 4.4. In particular, we randomly initialize the layers instead of using a Gaussian distribution or zero initialization. We observe a gradient explosion with the current learning rate. When we lower the learning rate (0.00001), the model trains over a longer period of time, while there are no discernible changes in the final results.

\section{Conclusion}
\label{sec:conclusion}
We propose MotionFlow as a novel normalizing flows approach that autoregressively conditions the output
distributions on the spatio-temporal input features. 
We apply our method to different tasks, including
trajectory prediction, motion prediction, time
series forecasting, and semantic segmentation.  
Our extensive experiments on 10 datasets across various tasks demonstrate that MotionFlow achieves state-of-the-art or competitive results in every case.
We believe that both deterministic and stochastic hidden representations are required to enable more trustworthy long-term predictions in comparison to fully stochastic or fully deterministic approaches. In future work, we intend to explore this relationship for more sophisticated tasks, such as video prediction.

\ifCLASSOPTIONcompsoc
\else
\fi

\section*{Acknowledgment}
The authors would like to thank Geotab Inc., the Natural Sciences and Engineering Research Council of Canada (NSERC), and Ingenuity Labs for their support of this work.

\ifCLASSOPTIONcaptionsoff
  \newpage
\fi



\bibliographystyle{IEEEtran}
\bibliography{ref}
%



%
\newpage

\begin{IEEEbiography}[{\includegraphics[width=1in,height=1.25in,clip,keepaspectratio]{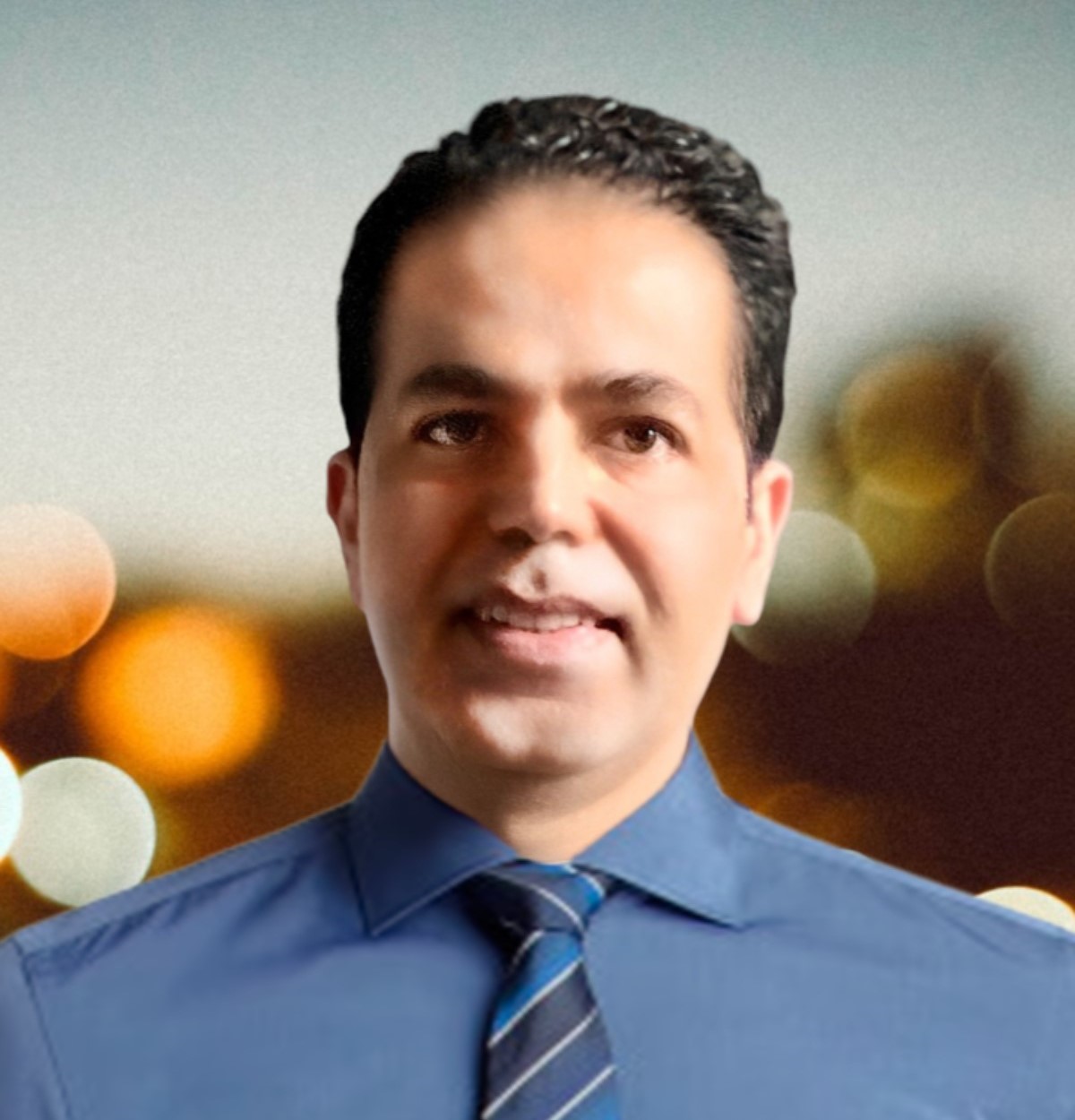}}]{Mohsen Zand} (Member, IEEE)
is a postdoctoral fellow who has been affiliated with the Robotics and Computer Vision (RCV) lab since January 2019. Additionally, he has been working with the Ambient Intelligence and Interactive Machines (Aiim) Lab, and Ingenuity Labs Research Institute, Queen’s University, Kingston, ON, Canada since 2020. Prior to joining the RCV Lab, Dr. Zand served as an Assistant Professor in the Computer Engineering Department at Azad University in Iran. He has been a member of the IEEE Computer Society since 2014. He was involved in different projects, such as object detection, motion prediction, medical imaging, moving object detection for remote sensing applications, and 3D object detection and recognition for the intelligent industrial robotics. His research interests are in the broad areas of computer vision, machine learning, multimedia content analysis, indexing and retrieval, and pattern recognition.
\end{IEEEbiography}

\begin{IEEEbiography}[{\includegraphics[width=1in,height=1.25in,clip,keepaspectratio]{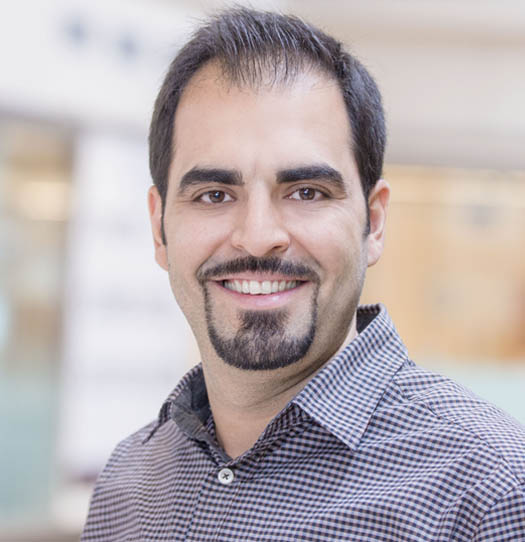}}]{Ali Etemad} (Senior Member, IEEE) is with the Department of Electrical and Computer Engineering, and Ingenuity Labs Research Institute, at Queen’s University in Canada. He holds the rank of Associate Professor as well as Mitchell Professor in AI for Human Sensing and Understanding. He leads the Ambient Intelligence and Interactive Machines (Aiim) lab, where his main area of research is machine learning and deep learning focused on human-centered applications with wearables, smart devices, and smart environments. Prior to joining Queen’s, he worked in the industry for several years as a data scientist. His works have appeared in top-tier venues such as CVPR, AAAI, ICCV, ECCV, ACM CHI, T-PAMI, T-AFFC, T-IP, and others. He is a co-inventor of 9 patents and has given many invited talks at different venues. Ali is an Associate Editor for IEEE Transactions on Artificial Intelligence, and has been a PC member/reviewer for many notable conferences and journals including NeurIPS, ICML, AAAI, CVPR, ICLR, ICCV, ECCV, ACII (senior PC), ICASSP, ISWC, ICMI, and others. He was the General Chair for the AAAI Workshop on Representation Learning for 
Responsible Human-Centric AI (2023), AAAI Workshop on Human-Centric Self-Supervised Learning (2022), Publicity Co-Chair for European Workshop on Visual Information Processing (2022), and Industry Relations Chair for Canadian Conference on AI (2019). Ali’s lab and research program have been funded by the Natural Sciences and Engineering Research Council (NSERC) of Canada, Ontario Centre of Innovation (OCI), Canadian Foundation for Innovation (CFI), and other organizations, as well as the private sector.
\end{IEEEbiography}


\begin{IEEEbiography}[{\includegraphics[width=1in,height=1.25in,clip,keepaspectratio]{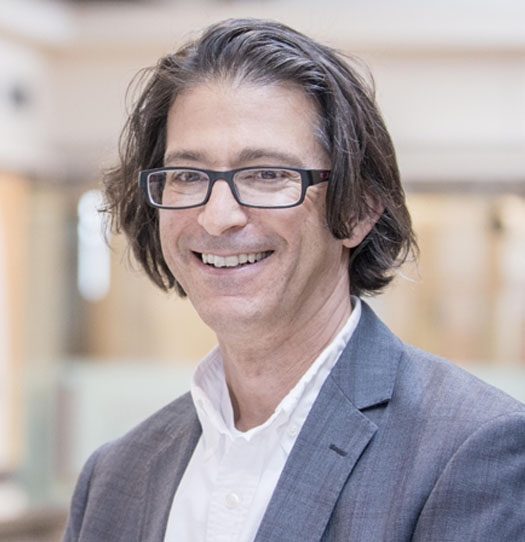}}]{Michael Greenspan} (Member, IEEE)
is a Professor in the Department of Electrical and Computer Engineering at Queen’s University, Kingston, Canada,
which he joined in 2001. Dr. Greenspan’s research investigates problems of computer vision and robotics, with a focus on the development of efficient and robust pose determination, object recognition and tracking methods using 3D data. Dr. Greenspan 
directs the Robotics and Computer Vision Lab (RCVLab) at Queen's,
and is a founding member of
Ingenuity Labs. He has over 100 refereed publications, including five patents, and he has served on the technical and program committees of over 60 international conferences in the fields of computer vision and robotics. Dr. Greenspan has led and participated in a number of collaborative projects with industry. In 2015-16, he was an Academic Resident at Seiko Epson Corporation, and in 2006-07 he served as Visiting Professor at University of Coimbra, Portugal. Greenspan holds membership in the Professional Engineers of Ontario, the IEEE Computer Society, and the IEEE Robotics and Automation Society. He has been the recipient of the Premier’s Research Excellence Award, the Canadian Image Processing and Pattern Recognition Society Young Investigator’s Award, and a number of Best Paper and Favorite Professor Awards. 
\end{IEEEbiography}




\end{document}